\documentclass[final]{cvpr}

\usepackage{times}
\usepackage{epsfig}
\usepackage{graphicx}
\usepackage{amsmath}
\usepackage{amssymb}

\usepackage{booktabs}
\usepackage{amsfonts}
\usepackage{amsmath}
\usepackage[numbers,sort]{natbib}
\usepackage{enumitem}
\usepackage{multirow}
\usepackage{balance}
\usepackage{tabularx,colortbl}

\usepackage[ruled,linesnumbered]{algorithm2e}
\usepackage{appendix}
\usepackage[pagebackref=true,breaklinks=true,colorlinks,bookmarks=false]{hyperref}

\def\cvprPaperID{2838} 
\def\confYear{CVPR 2021}

\graphicspath{{img/}}

\SetKwInput{KwInput}{Input}
\SetKwInput{KwOutput}{Output}
\SetKwInput{KwParameter}{Parameter}

\begin{document}

\title{Transformation Driven Visual Reasoning}


\author{%
    Xin Hong\textsuperscript{1,2} ~~%
    Yanyan Lan\textsuperscript{3,}\thanks{Corresponding author.} ~~%
    Liang Pang\textsuperscript{1,2} ~~%
    Jiafeng Guo\textsuperscript{1,2} ~~%
    Xueqi Cheng\textsuperscript{1,2} ~~%
    \\
    {\textsuperscript{1} CAS Key Laboratory of Network Data Science and Technology,} \\{Institute of Computing Technology, Chinese Academy of Sciences, Beijing, China} \\
    {\textsuperscript{2} University of Chinese Academy of Sciences, Beijing, China} \\
    {\textsuperscript{3} Institute for AI Industry Research, Tsinghua University, Beijing, China} \\
 {\tt\small \{hongxin19b, pangliang, guojiafeng, cxq\}@ict.ac.cn ~~lanyanyan@tsinghua.edu.cn}
}

\maketitle

\begin{abstract}
    This paper defines a new visual reasoning paradigm by introducing an important factor, i.e.~transformation. The motivation comes from the fact that most existing visual reasoning tasks, such as CLEVR in VQA, are solely defined to test how well the machine understands the concepts and relations within static settings, like one image. We argue that this kind of \textbf{state driven visual reasoning} approach has limitations in reflecting whether the machine has the ability to infer the dynamics between different states, which has been shown as important as state-level reasoning for human cognition in Piaget's theory. To tackle this problem, we propose a novel \textbf{transformation driven visual reasoning} task. Given both the initial and final states, the target is to infer the corresponding single-step or multi-step transformation, represented as a triplet (object, attribute, value) or a sequence of triplets, respectively. Following this definition, a new dataset namely TRANCE is constructed on the basis of CLEVR, including three levels of settings, i.e.~Basic (single-step transformation), Event (multi-step transformation), and View (multi-step transformation with variant views). Experimental results show that the state-of-the-art visual reasoning models perform well on Basic, but are still far from human-level intelligence on Event and View. We believe the proposed new paradigm will boost the development of machine visual reasoning. More advanced methods and real data need to be investigated in this direction. The resource of TVR is available at \url{https://hongxin2019.github.io/TVR}.
\end{abstract}

\section{Introduction}

Visual reasoning is the process of solving problems on the basis of analyzing the visual information, which goes well beyond object recognition~\citep{krizhevsky2012imagenet, ren2015faster, redmon2016you, he2017mask}. Though the task is easy for the human, it is tremendously difficult for vision systems, because it usually requires higher-order cognition and reasoning about the world. In recent years, several visual reasoning tasks have been proposed and attract a lot of attention in the community of computer vision, machine learning, and artificial intelligence. For example, the most representative visual question answering (VQA) tasks, such as CLEVR~\cite{johnson2017clevr}, define a question answering paradigm to test whether machines have spatial, relational, and other reasoning abilities for a given image. Visual entailment tasks such as NLVR~\citep{suhr2017corpus, suhr2019corpus} ask models to determine whether a sentence is true about the states of two images. Visual commonsense reasoning tasks, such as VCR~\cite{zellers2019vcr}, further require the model to provide a rationale explaining why its answer is right.

\begin{figure}[t]
    \begin{center}
        {\includegraphics[width=\columnwidth]{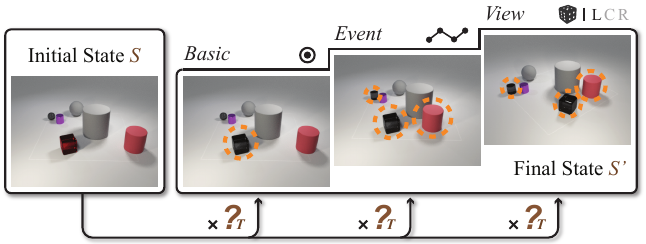}}
    \end{center}
    \caption{Illustration of three settings in TRANCE. \textbf{Basic:} Find the single-step transformation between the initial and final state. \textbf{Event:} Find the multi-step transformation between two states. \textbf{View:} Similar to Event but the view angle of the final state is randomly chosen from \textit{Left}, \textit{Center} (Default), and \textit{Right}.}
    \label{fig:tasks}
\end{figure}

We can see that these visual reasoning tasks are all defined at {\em state} level. For example, the language descriptions in NLVR as well as the questions and answers in VQA and VCR are just related to the concepts or relations within states, i.e.~an image, or two images. We argue that this kind of {\em state driven visual reasoning} fails to test the ability of reasoning dynamics between different states. Take two images as an example. In the first image, there is a cat on a tree, and in the second image, the same cat is under the tree. It is natural for a human to reason that the cat jumps down the tree after analyzing the two images. Piaget's cognitive development theory~\cite{piaget1977role} describes the dynamics between states as transformation, and tells that human intelligence must have functions to represent both the transformational and static aspects of reality. In addition, transformation is the key to tackle some more complicated tasks such as storytelling~\cite{huang2016visual} and visual commonsense inference~\cite{park2020visualcomet}. Though these tasks are closer to reality, they are too complicated to serve as a good testbed for transformation based reasoning. Because many other factors like representation and recognition accuracy may have some effects on the performance. Therefore, it is crucial to define a specific task to test the transformation reasoning ability.

In this paper, we define a novel {\em transformation driven visual reasoning } (TVR) task. Given the initial and final states, like two images, the goal is to infer the corresponding single-step or multi-step transformation. Without loss of generality, in this paper, transformations indicate changes of object attributes, so a single-step and multi-step transformation are represented as a triplet {\em (object, attribute, value)} and a sequence of triplets, respectively.

Following the definition of TVR, we construct a new dataset called TRANCE (\underline{Tran}sformation on \underline{C}L\underline{E}VR), to test and analyze how well machines can understand the transformation. TRANCE is a synthetic dataset based on CLEVR \cite{johnson2017clevr}, since it is better to first study TVR in a simple setting and then move to more complex real scenarios, just like people first study VQA on CLEVR and then generalize to more complicated settings like GQA. CLEVR has defined five types of attributes, i.e.~color, shape, size, material, and position. Therefore, it is convenient to define the transformation for each attribute, e.g.~the color of an object is changed from red to blue. Given the initial and final states, i.e.~two images, where the final state is obtained by applying a single-step or multi-step transformation on the initial state, a learner is required to well infer such transformation. To facilitate the test for different reasoning levels, we design three settings, i.e.~Basic, Event, and View. Basic is designed for testing single-step transformation. Event and View are designed for more complicated multi-step transformation reasoning, where the difference is that View further considers variant views in the final state. Figure~\ref{fig:tasks} gives an example of three different settings.

In the experiments, we would like to test how well existing reasoning techniques~\cite{johnson2017inferring, hudson2018compositional} work on this new task. However, since these models are mainly designed for existing reasoning tasks, they cannot be directly applied to TRANCE. To tackle this problem, we propose a new encoder-decoder framework named TranceNet, specifically for TVR. With TranceNet, existing techniques can be conveniently adapted to TVR. We test several different encoders, e.g.~ResNet~\cite{he2016deep}, Bilinear-CNN~\cite{lin2015bilinear} and DUDA~\cite{Park_2019_ICCV}. While for the decoder, an adapted GRU~\cite{cho2014learning} network is used to employ the image features and additional object attributes from TRANCE to predict the transformation, which is a sequence of triplets. Experimental results show that deep models perform well on Basic, but are far from human's level on Event and View, demonstrating high research potentials in this direction.

In summary, the contributions of our work include: 1) the definition of a new visual reasoning paradigm, to learn the dynamics between different states, i.e.~transformation; 2) the proposal of a new dataset called TRANCE, to test three levels of transformation reasoning, i.e.~Basic, Event, and View; 3) experimental studies of the existing SOTA reasoning techniques on TRANCE show the challenges of the TVR and some insights for future model designs.

\section{Related Works}

The most popular visual reasoning task is VQA. Questions in the earliest VQA dataset~\citep{VQA, balanced_binary_vqa, goyal2017making} are usually concerned about the category or attribute of objects. Recent VQA datasets have improved the requirements on image understanding by asking more complex questions, e.g.~Visual7W~\cite{zhu2016visual7w}, CLEVR~\cite{johnson2017clevr}, OK-VQA~\cite{marino2019ok}, and GQA~\cite{Hudson_2019_CVPR}. There are two other forms of visual reasoning tasks that need to be mentioned. Visual entailment tasks, such as NLVR~\citep{suhr2017corpus, suhr2019corpus}, and SNLI-VE~\citep{xie2018visual, xie2019visual}, ask models to determine whether a given description is true about a visual input. Visual commonsense reasoning~\cite{wang2018fvqa, zellers2019vcr} tasks require to use commonsense knowledge~\cite{talmor2019commonsenseqa} to answer questions. It is meaningful to solve these tasks which require various reasoning abilities. However, all the above tasks are defined to reason within a single state, which ignore the dynamics between different states.

Recently, several visual reasoning tasks have been proposed to consider more than one state. For example, CATER~\cite{girdhar2020cater} tests the ability to recognize compositions of object movements. While our target is to evaluate transformations, and our data contains more diverse transformations rather than just moving. Furthermore, CATER along with other video reasoning tasks such as CLEVRER~\cite{yi2020clevrer} and physical reasoning tasks~\citep{bakhtin2019phyre, Baradel_2020_ICLR_cophy} are usually based on dense states, which make the transformations hard to define and evaluate. Before moving to these complex scenarios, our TVR provides a simpler formulation by explicitly defining the transformations between two states, which is more suitable for testing the ability of transformation reasoning. CLEVR-Change~\cite{Park_2019_ICCV} is the most relevant work, which requires to caption the change between two images. The novelty is that TVR isolates the ability to reason about state-transition dynamics and supports a more thorough evaluation than captioning. Furthermore, CLEVR-Change only focuses on single-step transformations.

The concept of transformation has also been mentioned in many other fields. In~\citep{isola2015discovering, nagarajan2018attributes, li2020symmetry}, transformations are used to learn good attribute representations to improve the classification accuracy. In~\citep{fathi2013modeling, wang2016actions, liu2017jointly, alayrac2017joint, zhuo2019explainable}, object or environment transformations are detected to improve the performance of action recognition. However, those works in attribute learning and action recognition fields only consider single step transformation, thus not appropriate for testing a complete transformation reasoning ability. Some people may feel that procedure planning~\cite{chang2019procedure} has a similar task formulation to TVR and they are closer to reality. However, actions in planning data are usually very sparse. More importantly, both high-quality recognition and transformation reasoning are crucial to well model the task. As they are coupled together, it is not appropriate to use such tasks to evaluate the ability of transformation reasoning. That is why in this paper we use a more simple yet effective way to define the TVR task.

\section {The Definition of TVR} \label{sec:task}

TVR (Transformation driven Visual Reasoning) is a visual reasoning task that aims at testing the ability to reason the dynamics between states. Formally, we denote the state and transformation space as $\mathcal{S}$ and $\mathcal{T}$ respectively. The process of transforming the initial state into the final state can be illustrated as a function $f:\mathcal{S}\times\mathcal{T}\to\mathcal{S}$. Therefore, the task of TVR can be defined as inferring the transformation $T\in\mathcal{T}$ given both the initial state $S\in\mathcal{S}$ and final state $S'\in\mathcal{S}$. The space of transformation is usually very large, e.g.~any changes of pixel value can be treated as a transformation. Therefore, without loss of generality, we define the atomic transformation as an attribute-level change of an object, represented as a triplet $t=(o, a, v)$, which means the object $o$ with the attribute $a$ is changed to the value $v$. For example, the color of an object is changed to blue. And further taking the order of atomic transformations into account, the transformation can be formalized as a sequence of atomic transformations, denotes as $T=\{t_1, t_2, \dots, t_n\}$, where $n$ is the number of atomic transformations.

We make a distinction between the {\em single-step} ($n=1$) and {\em multi-step} ($n \ge 1$) transformation setting because they can be evaluated in different ways to reflect different levels of transformation reasoning abilities. For single-step problems, we can directly compare the prediction $\hat{T}$ with the ground-truth $T$ to obtain overall accuracy, as well as the fine-grained accuracy of each element in the triplet $(o, a, v)$. With these metrics, it is convenient to know how well a learner understands atomic transformations and to analyze why a learner performs not well. However, multi-step transformation problems cannot be evaluated in this way. This is because atomic transformations sometimes are independent so that the order of them can be varying, which makes the answer not unique anymore. For example, the procedures of cooking can not be disrupted but changing the order of cooking different dishes makes no difference to the final result. To tackle this multi-solution problem, we consider the reconstruction error as the evaluation metric. Specifically, the predicting transformation $\hat{T}$ is first applied to the initial state $S$ to obtain the predicted final state $\hat{S'}$. Then the transformed $\hat{S'}$ is compared with the ground-truth final state $S'$ to decide whether the predicting transformation is correct and how far is the prediction from a correct transformation. Therefore, the evaluation of multi-step problems focuses more on the ability to find all atomic transformations and a feasible order to arrange them.

With this definition, most existing state driven visual reasoning tasks can be extended to the corresponding transformation driven ones. For example, the VQA task, such as CLEVR, can be extended to ask the transformation between two given images, with answers as the required transformation. In the extension of NLVR, the task becomes to determine whether a sentence describing the transformation is true about the two images, e.g.~the color of the bus is changed to red. Since TVR itself is defined as an interpretation task, we do not need any further rational explanations, and the extension of VCR will stay the same as NLVR. We can see that the intrinsic reasoning target of these tasks is the same, that is to infer the correct transformation. While the difference lies in the manifestation.

\section{The TRANCE Dataset} \label{sec:dataset}

In this paper, we extend CLEVR by asking a uniform question, i.e.~what is the transformation between two given images, to test the ability of transformation reasoning. This section introduces how the TRANCE (\underline{Tran}sformation on \underline{C}L\underline{E}VR) dataset built following the definition of TVR.

\subsection{Dataset Setups}

We choose CLEVR~\cite{johnson2017clevr} to extend because CLEVR defines multiple object attributes, which can be changed conveniently. With the powerful Blender~\cite{blender2016blender} engine used by CLEVR, we are able to collect over 0.5 million samples with only computational costs.

\begin{table}[t]
    \centering
    \centerline{\includegraphics[width=0.9\columnwidth]{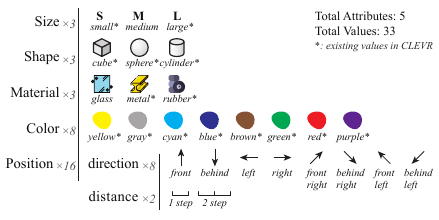}}
    \caption{Attributes and values in TRANCE.}
    \label{tab:values}
\end{table}

According to our definition, an atomic transformation is represented as a triplet $(o, a, v)$. In the following, we introduce attribute, value, and object one by one to demonstrate how to ground these factors.

The setup of the attribute is exactly the same as CLEVR. There are five attributes for each object, i.e.~size, color, shape, material, and position.

\begin{figure}[t]
    \centering
    \centerline{\includegraphics[width=\linewidth]{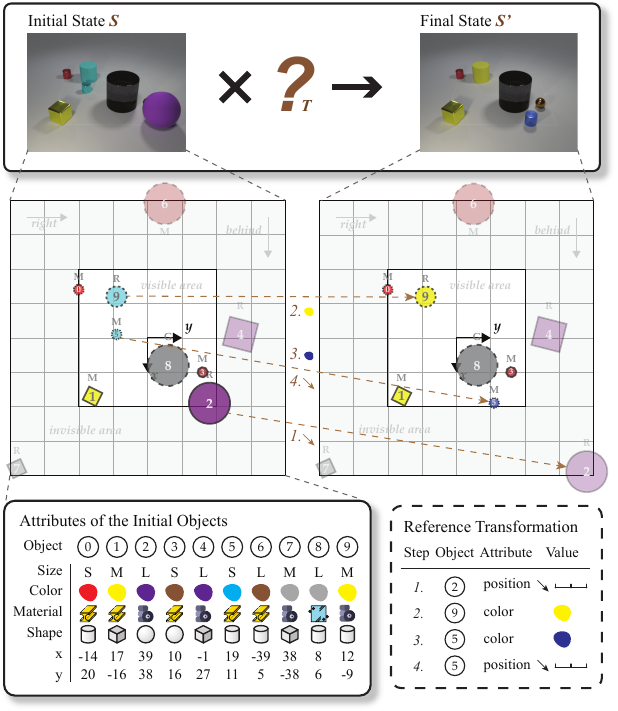}}
    \caption{An example from the Event setting.}
    \label{fig:example}
\end{figure}

The value in an atomic transformation is defined corresponding to the concerned attribute, as shown in Table~\ref{tab:values}. The values of the attributes except for position are similar to the default setting of CLEVR. Medium size and glass material are added to enrich the values. The value of position is carefully designed since it can be infinite in the space $\mathbb{R}^2$. To reduce the computation, we replace the absolute value of the position with a relative one by using direction and step to represent the value of a position transformation. We consider eight values for the variable direction, as shown in the Table~\ref{tab:values}. Then we define a coordinate system, in which $x$ and $y$ are both restricted to $[-40, 40]$, and objects can only be placed on integer coordinates. The variable step can be valued as 1 or 2, where 1 step equals 10 in our coordinate system. Except for normal moving action, we are also interested in whether the vision system could understand actions like moving in and moving out, so the plane is split into the visible area and the invisible area as shown in the middle two images of Figure~\ref{fig:example}, and the moving in and out operations can be defined correspondingly. To be reasonable, objects shouldn't be overlapped and moved out of the plane during transformation.

The setup of the object is basically the same as CLEVR. The only problem is how to represent an object in the answer. Existing methods such as CLEVR and CLEVR-Change use text which has ambiguity issues making the evaluation unreliable, while CLEVR-Ref+~\cite{liu2019clevr} employs bounding boxes which is specific but requires the additional ability of detection. Therefore, we propose a simple method that is specific and easy to evaluate by providing the attributes of the initial objects, as shown in Figure~\ref{fig:example}. In this way, an object can be referred to with the assigned number. Note machines still need to perform their own recognition to align objects in images with given attributes.

To generate TRANCE, the first step is the same as CLEVR, which is randomly sampling a scene graph. According to the scene graph, CLEVR then generates questions and answers with a functional program and renders the image with Blender. Different from CLEVR, the next step in TRANCE generation becomes randomly sampling a sequence of atomic transformations ($n \in \{1,2,3,4\} $), which is called as the reference transformation, to transform the initial scene graph to the final scene graph. At last, two scene graphs are rendered into images ($h:240\times w:320$). The attributes of the initial objects can be easily obtained from the initial scene graph.

To reduce the potential bias in TRANCE, we carefully control the sampling process of scene graph and transformation by balancing several factors. In scene graph sampling, we balance objects' attributes and the number of visual objects in the initial state. In transformation sampling, the length of the transformation, the object number, n-gram atomic transformation, and the move type are all balanced. Throughout all elements, N-gram atomic transformation is the hardest to be balanced and it refers to the sub-sequence of atomic transformations with the length of $n$. By balancing these factors, we reduce the possibility that a learner utilizes statistics features in the data to predict answers. In the supplementary material, we show the statistics of the dataset and our balancing method in detail.

\subsection{Three Levels of Settings} \label{sec:settings}
To facilitate the study on different levels of transformation reasoning, we design three settings, i.e.~Basic, Event, and View. Basic is designed for single-step transformation and Event is for multi-step transformation. To further evaluate the ability of reasoning transformation under a more real condition, we extend Event with variant views to propose View. Figure \ref{fig:tasks} compares three different settings, more examples can be found in the supplementary material.

\textbf{Basic.}
Basic is set as the first simple problem to mainly test how well a learner understands atomic transformations. The target of Basic is to infer the single-step transformation between the initial and final states. That is, given a pair of images, the task is to find out which attribute $a$ of which object $o$ has been changed to which value $v$. We can see that this task is similar to the previous game `Spot the Difference'~\cite{jin2013spoid}, in which the player is asked to point out the differences between two images. However, Basic is substantially different from the game. The game focuses on the pixel level differences while Basic cares about the object level differences. Therefore, Basic can be viewed as a more advanced visual reasoning task than the game.

\textbf{Event.}
It is obviously not enough to consider only the single-step transformation. In reality, it is very common that multi-step transformation exists between two states. Therefore, we construct this multi-step transformation setting to test whether machines can handle this situation. The number of transformations between the two states is randomly set from 1 to 4. The goal is to predict a sequence of atomic transformations that could reproduce the same final state from the initial state. To resolve this problem, a learner must find all atomic transformations and arrange them with a feasible order. Compared with Basic, it is possible to have multiple transformations, which improves the difficulty of finding them all. Meanwhile, the order is essential in the Event setting because atomic transformations may be dependent. For example, in Figure \ref{fig:example}, two moving steps, i.e.~1st step and 4th step, cannot be exchanged, otherwise, object 5 and object 2 will overlap.

\textbf{View}
The view angle of Basic and Event is fixed, which is not the case in real applications. To tackle this problem, we extend the Event setting to View, by capturing two states with cameras in different positions. In practice, for simplicity but without loss of generality, we set three cameras, placed in the left, center, and right side of the plane. The initial state is always captured by the center camera, while for the final state, images are captured with all three cameras. Thus, for each sample, we obtain three pairs for training, validation, and testing with the same initial state but different views of final states. In this way, we are capable to test whether a vision system can understand object-level transformation with variant views.

\subsection{Evaluation Metrics} \label{sec:metrics}

For the single-step transformation setting, i.e.~Basic, the answer is unique. Therefore, we can evaluate the performance by directly comparing the prediction with the reference transformation, which is also the ground-truth transformation. Specifically in this paper, we consider two types of accuracy. The first one is fine-grained accuracy corresponds to three elements in transformation triplet, including object accuracy, attribute accuracy, and value accuracy, denoted as {\em ObjAcc}, {\em AttrAcc}, and {\em ValAcc} respectively. The other one is the overall accuracy, which only counts the absolutely correct transformation triplets, denoted as {\em Acc}.

For multi-step transformation settings, i.e~Event and View, it is not suitable to use the above evaluation metrics, because the answers may not be unique. Therefore, the predicting atomic transformation sequence is evaluated by checking whether it could reproduce the same final state as the reference transformation. Specifically, we first obtain the corresponding final state $\hat{S}'$ by applying the predicting transformation $\hat{T}$ to the initial state $S$, i.e.~$S \times \hat{T} \rightarrow \hat{S'}$. Then a {\em distance} is computed by counting the attribute level difference between $\hat{S'}$ and the ground-truth final state $S'$. To eliminate the influence of different transformation lengths on distance, we normalize it by the length of the reference transformation to get a {\em normalized distance}. Averaging these two metrics on all samples, we obtain {\em AD} and {\em AND}. We further consider the overall accuracy denoted as {\em Acc} when the distance equals to zero. In addition, we are interested to see without considering the order, whether all atomic transformations are found, by omitting all constraints such as no overlapping to compute the loose accuracy, which denotes as {\em LAcc}. At last, to measure the ability of assigning the right order when all atomic transformations have been found, the error of order $EO = (LAcc - Acc) / LAcc$ is computed. In summary, five evaluation metrics are used in multi-step transformation settings, i.e.~{\em AD}, {\em AND}, {\em LAcc}, {\em Acc}, and {\em EO}.

In the evaluation of multi-step transformation problems, an important step is to obtain the predicting final state $\hat{S}'$ by applying the predicting transformation $\hat{T}$ to the initial state $S$. This function has already been implemented in the previously mentioned data generation system and we reuse it in our multi-step transformation evaluation system. Except for the usage of evaluation, this evaluation system can also be used to generate signals as rewards for reinforcement learning, which is explored in Section \ref{sec:main_res}.

\section{Experiments}

In this section, we show our experimental results on the three settings of TRANCE, i.e. Basic, Event, and View. We also conduct analyses to show some insights about machines' ability of reasoning transformation.

\subsection{Models} \label{sec:model}

\begin{figure}[t]
    \centering
    \centerline{\includegraphics[width=0.9\linewidth]{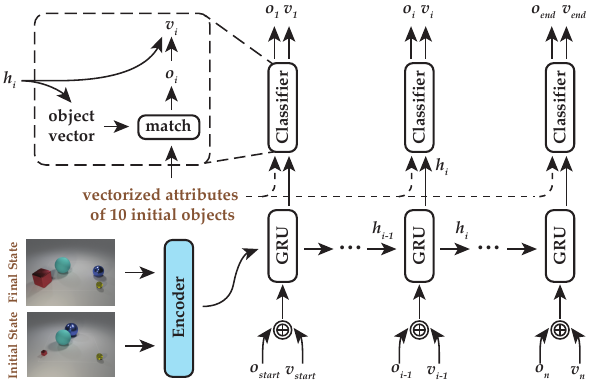}}
    \caption{The architecture of TranceNet.}
    \label{fig:network}
\end{figure}

Firstly, we would like to test how well existing methods work on this new task. However, since the inputs and outputs of TVR are quite different from existing visual reasoning tasks, existing methods like~\cite{johnson2017inferring, hudson2018compositional} cannot be directly applied. So we design a new encoder-decoder style framework named TranceNet. As Figure \ref{fig:network} shows, start from the input image pairs, an encoder first extracts features, and then a GRU based decoder is employed to generate transformation sequences. The following of this section briefly introduces our TranceNet framework while the implementation details can be found in the supplementary material. To compare with the human, for each of the three settings, we also collect results of 100 samples in total. These results come from 10 CS Ph.D. candidates who are familiar with our problems and the testing system.

\textbf{Encoder.} The goal of an encoder is to extract effective features from image pairs, which are mainly associated with the difference between the two states. There are two ways to extract these features, namely single-stream way and two-stream way. Single-stream way directly inputs two images into a network to extract features, while two-stream way first separately extracts image features and then interacts them in feature-level. In this paper, we evaluate six methods. In terms of the single-stream way, we test two networks, i.e.~Vanilla \textbf{CNN} and \textbf{ResNet}~\cite{he2016deep}, combined with two preprocessing strategies, i.e.~subtraction ($-$) and concatenation ($\oplus$). For example, we use ResNet$_\oplus$ to represent a ResNet fed with concatenated image pairs. Another two methods are Bilinear CNN (\textbf{BCNN}~\cite{lin2015bilinear}) and a recently proposed method called \textbf{DUDA}~\cite{Park_2019_ICCV}, which operate as the two-stream way. BCNN is a classical model for fine-grained image classification to distinguish categories with small visual differences. DUDA is originally proposed for change detection and captioning. The main difference between BCNN and DUDA lies in the way of feature-level interaction.

\textbf{Decoder.} The decoder is used to output a feasible transformation sequence from the extracted image features. All six encoders share the same modified GRU~\cite{cho2014learning} network, which is a commonly used technique for sequence generation. As shown in Figure \ref{fig:network}, the major difference between our GRU network and a standard one lies in the additional classifiers. A classifier unit accepts attributes of the initial objects and the current hidden state from the GRU cell, and then outputs the object and value of the current step. In detail, at a certain step, an object vector is first computed from the hidden state. Then the object number is obtained by matching the object vector with the initial objects using cosine similarity. Finally, the most similar object vector from initial objects and the hidden state are used to predict the value. In TRANCE, attributes are implied by values, for example, blue indicates that the attribute is color, so that the output of a classifier does not explicitly include an attribute.

Since all these models share the same decoder, we denote these models by their encoders' names hereafter.

\textbf{Training.} The loss function of a single sample consists of two cross-entropy losses for object and value respectively, which can be represented as:
\begin{equation}
\begin{split}
\mathcal{L}(T, \hat{T}) = -\frac{1}{n} \sum_{i=1}^{n}  ( t_{i}^{o} \cdot \log\hat{t}_{i}^{o} + t_{i}^{v} \cdot \log\hat{t}_{i}^{v}),
\end{split}
\end{equation}
where $n$ is the transformation length of the sample, $t_i^{o}$ and $t_i^{v}$ denote the object and value in the $i$-th step of transformation. The training loss is the average of losses over all training examples. During training, we use teacher forcing~\cite{teacherforcing} for faster convergence in two parts of TranceNet. Firstly, at each step, we follow the practice in sequence learning such as machine translation by using the object and value from the given reference transformation as the inputs of the GRU unit. Additionally, in the classifier, objects from reference transformations are used to predict values.

\begin{table*}[ht]
    \setlength{\tabcolsep}{0.5em}
    \begin{center}
    \begin{small}
    \begin{tabular}{lrrrrrrrrrrrr}
    \toprule
         \multirow{2}{*}{Model} & \multicolumn{4}{c}{Basic} & \multicolumn{4}{c}{Event} & \multicolumn{4}{c}{View} \\
         \cmidrule[.5pt](lr){2-5}  \cmidrule[.5pt](lr){6-9}  \cmidrule[.5pt](lr){10-13}
         &   ObjAcc$\uparrow$ &   AttrAcc$\uparrow$ &   ValAcc$\uparrow$ &   Acc$\uparrow$ &   AD$\downarrow$ &   AND$\downarrow$ &   LAcc$\uparrow$ &   Acc$\uparrow$ &   AD$\downarrow$ &   AND$\downarrow$ &   LAcc$\uparrow$ &   Acc$\uparrow$ \\
         \midrule
    CNN$_{-}$         & 0.9584          & 0.9872          & 0.9666          & 0.9380          & 1.5475          & 0.5070          & 0.4587          & 0.4442          & 2.2376          & 0.8711          & 0.2344          & 0.2286           \\
    CNN$_{\oplus}$    & 0.9581          & 0.9889          & 0.9725          & 0.9420          & 1.4201          & 0.4658          & 0.4981          & 0.4838          & 2.2517          & 0.8764          & 0.2350          & 0.2285           \\
    ResNet$_{-}$      & 0.9830          & 0.9969          & \textbf{0.9935} & 0.9796          & 1.0974          & \textbf{0.3417} & 0.5972          & 0.5750          & \textbf{1.1068} & 0.3749          & 0.5484          & 0.5272           \\
    ResNet$_{\oplus}$ & \textbf{0.9852} & \textbf{0.9980} & 0.9928          & \textbf{0.9810} & \textbf{1.0958} & 0.3469          & \textbf{0.6019} & \textbf{0.5785} & 1.1148          & \textbf{0.3731} & \textbf{0.5525} & \textbf{0.5305}  \\
    BCNN              & 0.9705          & 0.9950          & 0.9788          & 0.9571          & 1.1081          & 0.3582          & 0.5746          & 0.5560          & 1.2633          & 0.4395          & 0.4977          & 0.4784           \\
    DUDA              & 0.9453          & 0.9888          & 0.9692          & 0.9320          & 1.5261          & 0.4975          & 0.5025          & 0.4856          & 1.5352          & 0.5242          & 0.4746          & 0.4590           \\
    \midrule
    Human               & 1.0000               & 1.0000               & 1.0000               & 1.0000  &  0.3700   &  0.1200     & 0.8300 & 0.8300 & 0.3200 & 0.0986 & 0.8433 & 0.8433  \\
        \bottomrule
        \end{tabular}
        \end{small}
    \end{center}
    \caption{Model and human performance on Basic, Event, and View.}
    \label{tab:results}
\end{table*}

\subsection{Results on Three Settings} \label{sec:main_res}

In this section, we first present our experimental results on the three settings of TRANCE, and then provide some in-depth analyses on the results.

Six models are tested in the Basic setting of TRANCE, i.e.~CNN$_-$, CNN$_\oplus$, ResNet$_-$, ResNet$_\oplus$, BCNN, DUDA. From the results in the left part of Table~\ref{tab:results}, we can see that all models perform quite well, in the sense that the performance gap between these models and the human is not very large. Comparing these models, both versions of ResNet perform better than BCNN and DUDA. As we mentioned before, CNN and ResNet are single-stream methods while BCNN and DUDA are two-stream methods. Since the model size of ResNet, BCNN, and DUDA is similar, we can conclude that the single-stream way is better than the two-stream way on the Basic setting. Further checking the fine-grained accuracy, we can see this gap comes from the ability to find the correct objects and values, while all models are good at distinguishing different attributes.

\begin{table}[t]
    \begin{center}
        \begin{small}
            \begin{tabular}{lrrrr}
            \toprule
             Model           & AD$\downarrow$              & AND$\downarrow$             & LAcc$\uparrow$            & Acc$\uparrow$                 \\
            \midrule
 ResNet$_{\oplus}$ & 1.0958          & 0.3469          & 0.6019          & 0.5785           \\
 \hspace{0.2em} + \textit{corr}             & 1.0579          & 0.3316          & 0.6215          & 0.5978           \\
 \hspace{0.2em} + \textit{dist} & 1.0528          & 0.3319          & 0.6180          & 0.5938           \\
 \hspace{0.2em} + \textit{corr \& dist}        & \textbf{1.0380} & \textbf{0.3251} & \textbf{0.6230} & \textbf{0.6001}  \\
            \bottomrule
            \end{tabular}
        \end{small}
    \end{center}
\caption{Results of ResNet$_{\oplus}$ trained using REINFORCE~\cite{williams1992simple} with different rewards on Event.}
\label{tab:rl}
\end{table}

The experimental results on Event are shown in the middle part of Table~\ref{tab:results}. We can see that this task is very challenging for machines, since there is an extremely big performance gap between models and the human. That is because the answer space rises exponentially when the number of steps increases. In our experiments, the size of answer space is $\sum_{i=1}^4 (33 \times 10)^i$, about 11.86 billion. The performance (e.g.~Acc) gap between CNN and ResNet models becomes larger from Basic to Event, which suggests larger encoders have advantages in extracting sufficient features to decode transformation sequences. 

We also employ reinforcement learning to train models. Specifically, the evaluation system introduced in Section \ref{sec:metrics} can provide signals include the \textit{correctness} of a prediction and the \textit{distance} of a prediction to the correct one. These signals are able to be used as rewards in REINFORCE~\cite{williams1992simple} algorithm to further train ResNet$_{\oplus}$ models. Table \ref{tab:rl} shows that all three rewards significantly improve the performance, and the difference among them is small.

The right part of Table~\ref{tab:results} shows the results on the View setting. While humans are insensitive to view variations, the performances of all deep models drop sharply from Event to View. Among these models, the most robust one is DUDA. From the fact that BCNN has a similar two-stream architecture but performs worse, we can see that DUDA's way of directly interacting two state features is more effective for tackling the view variation.

\subsection{Detailed Analysis on Event and View} \label{sec:analysis}

According to the above experimental results, the performances on Event and View are not good. So we conduct some detailed analysis to help understand the task and provide some insights for future model designs.

\begin{figure}[t]
    \centering
    \includegraphics[width=0.9\linewidth]{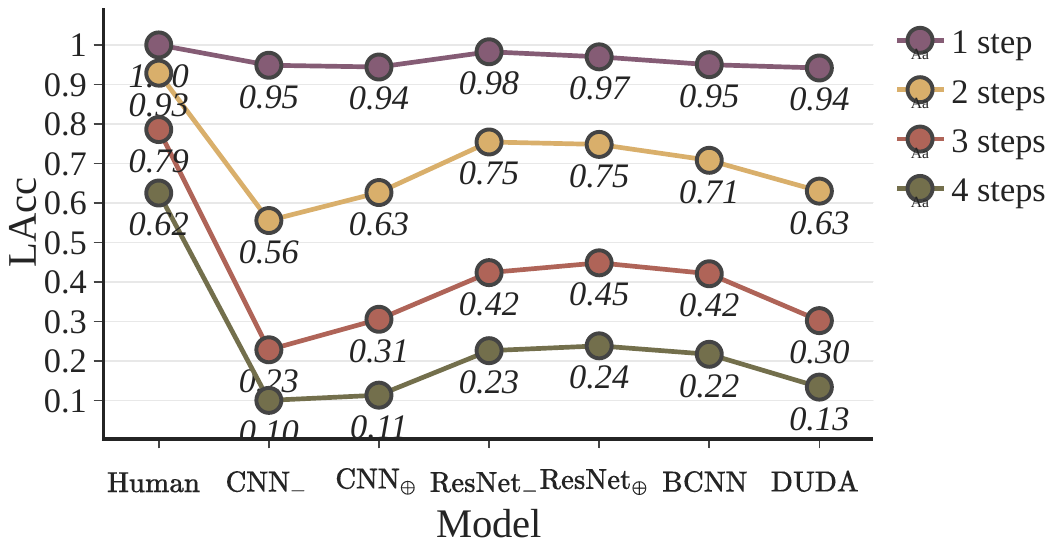}
    \caption{Results on Event with respect to different steps.}
    \label{fig:event}
\end{figure}

Firstly, we analyze the effect of transformation sequence length on Event, which is the main factor to make performance worse than Basic. Specifically, we separate all test samples into four groups based on their lengths, i.e.~samples with $k$-step transformation $(k=1,2,3,4)$. Then we plot the LAcc for each group in Figure~\ref{fig:event}. From the results, we can see that both human and deep models work quite well when the length is short, e.g.~1. As the length increases, humans still have the ability to well capture the complicated transformations. However, the deep models decline sharply. Take CNN$_{-}$ as an example, the performances for the four different groups are 95\%, 56\%, 23\%, and 10\%. These results indicate that future studies should focus more on how to tackle transformations with long steps.

\begin{table}[t]
\begin{center}
    \begin{small}
    \begin{tabular}{lrrr}
    \toprule                
     Encoder           & LAcc$\uparrow$            & Acc$\uparrow$             & EO$\downarrow$              \\
    \midrule
     Random (avg. of 100)        & 1.0000          & 0.4992          & 0.5008 \\
    \midrule
     CNN$_{-}$         & 0.2276          & 0.2067          & \textbf{0.0915} \\
     CNN$_{\oplus}$    & 0.2596          & 0.2244          & 0.1358          \\
     ResNet$_{-}$      & 0.3478          & 0.2997          & 0.1382          \\
     ResNet$_{\oplus}$ & \textbf{0.3862} & \textbf{0.3205} & 0.1701          \\
     BCNN              & 0.2949          & 0.2612          & 0.1141          \\
     DUDA              & 0.2500          & 0.2147          & 0.1410          \\
    \midrule
     Human               &  0.7273 & 0.7273 & 0.0000 \\
    \bottomrule
    \end{tabular}
    \end{small}
\end{center}
\caption{Results on 7.8\% order sensitive samples from Event.}
\label{tab:order}
\end{table}

Then we analyze the effect of the order on Event, which is another important factor in this data. According to our statistics, there are about 7.8\% test samples\footnote{We have also tried other data with different percentages, e.g.~25\%, and the results are similar.} exist certain permutations that violate our constraints such as no overlapping. That is to say, these samples are order sensitive. Even if a model is able to find all correct atomic transformations, the result still could be wrong without carefully considering an order to arrange them. Table~\ref{tab:order} shows the results on these order sensitive samples, where EO is directly defined to measure the influence of order, LAcc and Acc are just listed for reference. From the results, we can see that EO of the human is zero. Therefore, once humans find all correct atomic transformations, it is easy to figure out the orders meanwhile. However, for all deep models, the EOs are larger than zero, which indicates a clear effect of the order on the reasoning process. In order to find out the extent of the effect, i.e., whether $0.0915\sim0.1701$ means a large deviation, we perform an experiment on 100 randomly selected order sensitive samples. Specifically for each sample, we randomly assign an order for ground truth atomic transformations. As a result, the EO is 0.5008, which could be viewed as an upper bound of the order error. Therefore, the current deep models indeed have some ability to tackle the orders, but there still leaves some room for improvements.

\begin{figure}[t]
    \centering
    \includegraphics[width=0.9\linewidth]{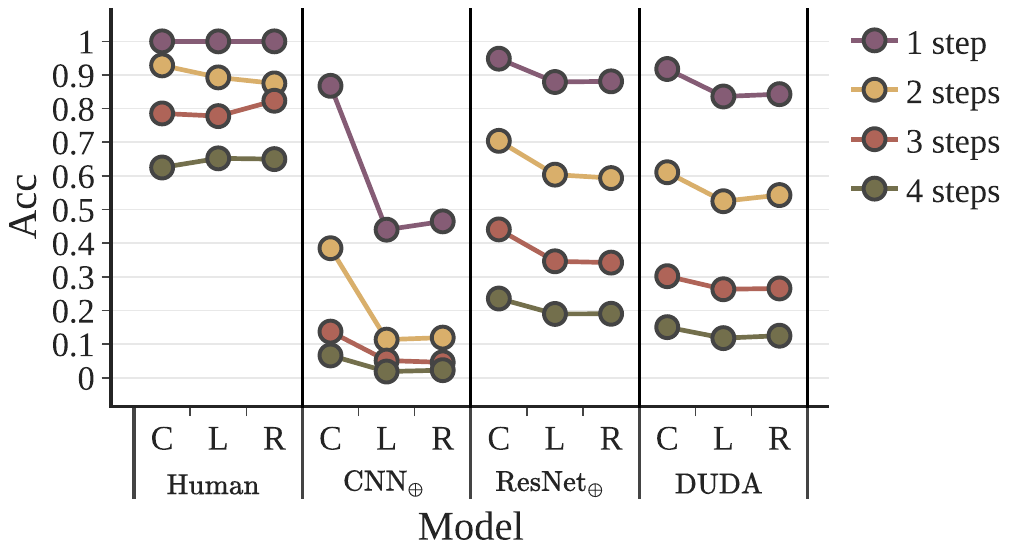}
    \caption{Results for different final views (\textbf{C}enter, \textbf{L}eft, \textbf{R}ight).}
    \label{fig:view_result}
\end{figure}

At last, we analyze the effect of view variation. For each model, we provide the results of different final views, as shown in Figure~\ref{fig:view_result}. Please note that the results of CNN$_{-}$, ResNet$_{-}$, and BCNN are quite similar to CNN$_{\oplus}$, ResNet$_{\oplus}$ and DUDA, so we just give the results from latter three typical models. Firstly, the results of humans across different views change small, demonstrating human's powerful ability of adapting to different views. In some cases, humans perform even better when views are changed than unchanged. That is because when the view is altered, humans usually spend more time solving the problems, which decreases the chances of making errors. Conversely, deep learning models share a similar trend that view variations will hurt the performance. Among these models, CNN decreases the most, while DUDA shows its robustness. In conclusion, models with more parameters are more robust to view variations and feature-based interaction like the way used in DUDA, is helpful. 

\subsection{Analysis of Training Data Size}

\begin{figure}[t]
    \centering
    \includegraphics[width=0.83\linewidth]{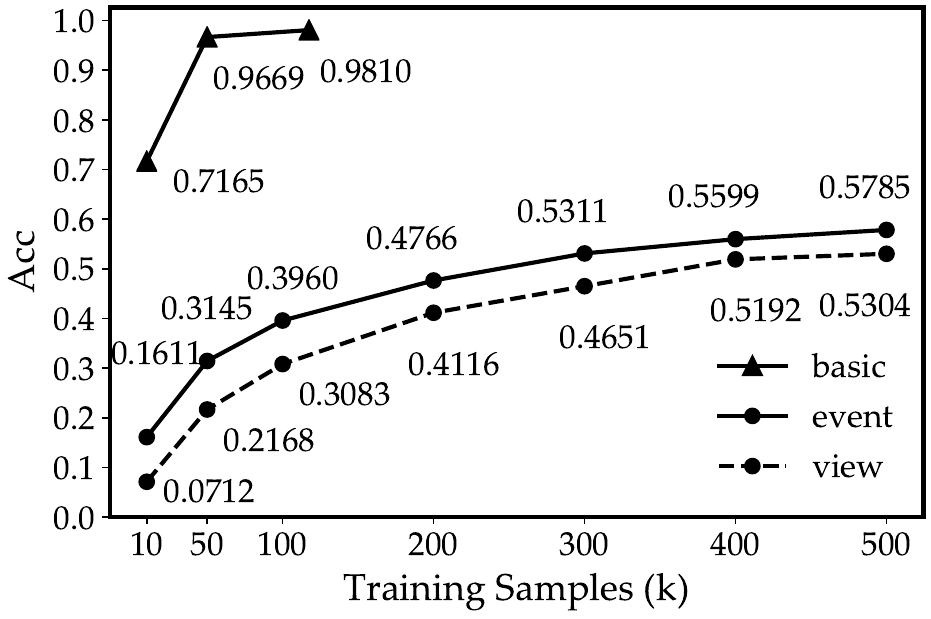}
    \caption{Results of ResNet$_{\oplus}$ with different training size.}
    \label{fig:size}
\end{figure}

In our experiments, we find that data size is an important factor for training and evaluation. Therefore, we use ResNet$_\oplus$ as an example to study the influence of this factor. From Figure \ref{fig:size}, we can see that more training samples bring significant benefits when the number is less than 50k on Basic and 200k on Event and View. After that, the benefits become smaller and smaller. Those results are consistent with the common knowledge that relatively large data is required to well train a deep model. These results also show TRANCE has sufficient size in our experiments.

\section{Conclusion}

To tackle the problem that most existing visual reasoning tasks are solely defined on static settings and cannot well capture the dynamics between states, we propose a new visual reasoning paradigm, namely transformation driven visual reasoning (TVR). Given the initial and final states, the target is to infer the corresponding single-step transformation or multi-step transformations, represented by a triplet (object, attribute, value) or a sequence of triplets, respectively. In this paper, as an example, we use CLEVR to construct a new synthetic data, namely TRANCE, which includes three different levels of settings, i.e.~Basic for single-step transformation, Event for multi-step transformation, and View for multi-step transformation with variant views. To study the effectiveness of existing SOTA reasoning techniques, we propose a new encoder-decoder framework named TranceNet and test six models under this framework. The experimental results show that our best model works well on Basic, while still has difficulties solving Event and View. Specifically, the difficult point of Event is to find all atomic transformations and arrange them with a feasible order, especially when the length of the sequence is large. While for View, the view variations bring great challenges to these models, but have little impact on humans. 

In the future, we plan to investigate from both model and data perspectives by testing methods like neural symbolic approaches and constructing a real dataset to study TVR.

\section*{Acknowledgement}

This work was supported by the National Natural Science Foundation of China (NSFC) under Grants No. 61773362, and 61906180, Beijing Academy of Artificial Intelligence (BAAI) under Grants BAAI2020ZJ0303, the National Key R\&D Program of China under Grants No. 2016QY02D0405, the Tencent AI Lab Rhino-Bird Focused Research Program (No. JR202033).

\appendix
\appendixpage

    This document aims at providing additional materials to supplement our main submission. We first show the detail of data balancing on TRANCE in Section \ref{sec:balance}. Then, we give more details on the implementation of the baseline models and training in Section \ref{sec:detail}. Next in Section \ref{sec:human_test}, we describe the test system we used for collecting results from humans. Finally in Section \ref{sec:examples}, we provide extra examples of three settings from TRANCE, i.e.~Basic, Event, and View.

\section{Dataset Balance} \label{sec:balance}

Data balancing is an important factor to be considered when constructing TRANCE. Several factors are balanced in TRANCE, so that a learner is expected to reason the transformation without utilizing the biased features such as the length of transformation in data. Without considering the image rendering, the data generation process consists of two stages, i.e.~sampling an initial scene graph and sampling a transformation sequence to transform the initial scene graph into the final scene graph. In the following of this section, we first introduce the factors that are balanced in these two stages and then describe the method we use.

\begin{figure*}[t]
    \centering
    \includegraphics[width=\linewidth]{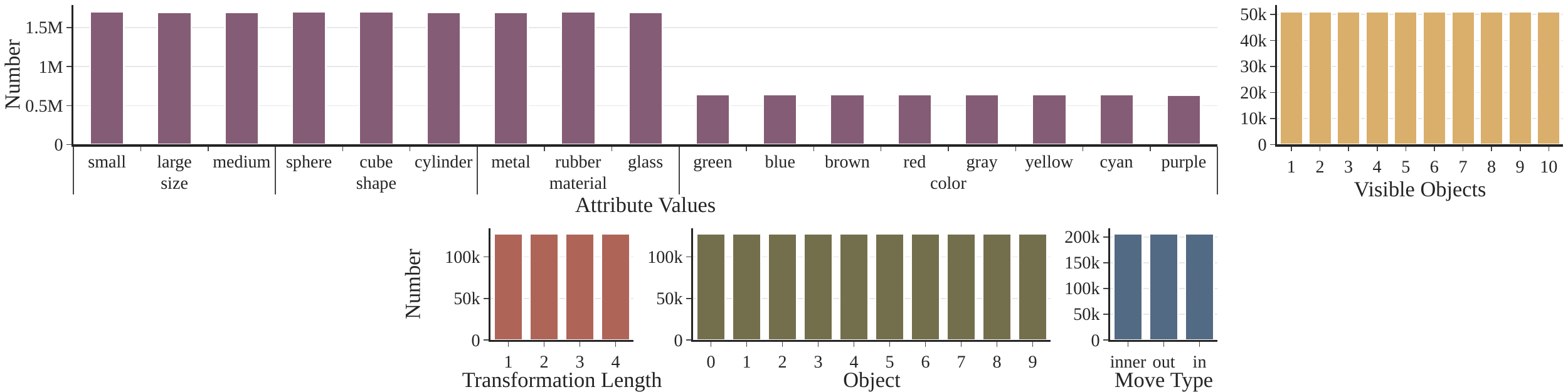}
    \caption{The statistics of balanced factors in the TRANCE dataset. \textbf{Top Row:} The attribute values and the visible objects which are balanced during sampling the initial scene graphs. \textbf{Bottom Row:} The transformation length, object number, and move type which are balanced when sampling transformation sequences.}
    \label{fig:statistics}
\end{figure*}

When sampling the initial scene graph, the attributes of all objects and the number of visible objects are balanced. Recall that the plane is separated into the visible area and invisible area and only objects in the visible area appear in the image of the initial state. The two diagrams on the top row of Figure~\ref{fig:statistics} show the statistics of these two factors. We can see that they are strictly balanced.

When sampling the transformation sequence, we balance four factors in total. The first factor is the length of transformation so that samples with different transformation lengths are equal in terms of size. The statistic result of the transformation length can be found in Figure~\ref{fig:statistics} on the left side of the second row. The other three factors are all about the elements of atomic transformations including the object number for object and the move type and $n$-gram atomic transformation for the value. The object number is directly balanced over all samples and the result is shown in the middle of the second row in Figure~\ref{fig:statistics}. The move type is also strictly balanced and the statistics is shown in the right of the second row in Figure~\ref{fig:statistics}. As for the $n$-gram atomic transformation, it should be handled carefully, since for a specific initial scene graph, the availability of different atomic transformations is different. For example, changing the color of one object can always be successful, but changing the position of an object with a specific direction and step may be failed because of overlapping. Thus, the concurrence of different atomic transformations has different probability. For example, four atomic transformations on position will be less possible than four atomic transformations on color exist in one sequence. Therefore, we need to consider to balance the value throughout the sub-sequence of atomic transformations. In the following, we will call a sub-sequence with the length $n$ as a $n$-gram atomic transformation. Table \ref{tab:n_gram} shows the statistics of this factor. For each $n$-gram, the number of different options to be chosen is shown in the first row. For example, we have 33 different values so that the options of 1-gram are 33 and that of 2-gram is $33^2=1089$ and so on. The process of counting $n$-gram atomic transformations can be regarded as counting the sub-sequences using a n-length sliding window with 1 stride on transformation sequences. For example, to count 2-gram atomic transformations on a 4-step transformation, we use a 2-length sliding window with 1 stride and there will be three 2-gram atomic transformations. The rows below the options in Table \ref{tab:n_gram} are calculated on the counting results. From the table, the standard variance is very small compared to the mean value, which means the concurrence of different atomic transformations is well balanced. It should be note that the size of TRANCE is 0.5 million, which is not enough to cover all 4-gram options, but the analysis of training data size in our experiments has proved our data is sufficient for training a deep model. In conclusion, the dataset is well balanced to eliminate the potential bias that are not related to the target of transformation reasoning.

The method we used to balance all the above factors is called balanced sampling. The basic idea of this method is changing the probability of the sampling targets dynamically according to previously generated samples. Algorithm~\ref{alg:sample} shows how to sample an option given the count table of previously generated options.

\begin{table}[t]
\begin{center}
\begin{small}
    \begin{tabular}{lrrrr}
    \toprule                
                   &   1-gram &  2-gram   & 3-gram &  4-gram \\
    \midrule
    options    & 33        & 1,089      & 35,937     & 1,185,921   \\
    \midrule
    min                 & 38,635     & 697       & 7         & 0         \\
    max                 & 38,638     & 708       & 15        & 3         \\
    median              & 38,636     & 703       & 11        & 0         \\
    mean                & 38,636     & 702.5     & 10.64     & 0.1075    \\
    std                 & 0.7714     & 2.2854    & 0.7880    & 0.3150    \\
    \bottomrule
    \end{tabular}
\end{small}
\end{center}
\caption{The statistics of the $n$-gram atomic transformations in TRANCE.}
\label{tab:n_gram}
\end{table}

\begin{algorithm}[t]
\KwInput{available $k$ options $O = \{o_1, o_2, ... , o_k\}$, corresponding count table $N = \{n_1, n_2, ... , n_k\}$;}
\KwOutput{sampled option $o_r$;}
\KwParameter{tolerance $t = 0.1$ (default);}
\BlankLine

$n_{max}$ = max($n_1, n_2, ... , n_k$) \;
$c_i = n_{max} - n_i + t$ \;
$p_i = \frac{c_i}{\sum_{i=1}^{k}{c_i}}$ \;

$o_r$ = randomly sample an option from $\{o_1, o_2, ... , o_k\}$ with probability $\{p_1, p_2, ... , p_k\}$ \;
 \caption{Balanced Sampling}
 \label{alg:sample}
\end{algorithm}

\section{Implementation Details} \label{sec:detail}

The code for data generation is rewritten on the basis of the CLEVR\footnote{https://github.com/facebookresearch/clevr-dataset-gen}. In terms of training, we use PyTorch~\cite{NEURIPS2019_9015} as our deep learning framework. All of the code can be found at our Github repository\footnote{https://github.com/hughplay/TVR}. In the following, we introduce the implementation of our baseline models and the training process in detail.

Table~\ref{tab:model} shows the architectural details of different baseline models under the TranceNet framework. In the encoder part, both CNN$_{-}$ and CNN$_{\oplus}$ use a 4-layer CNN as the backbone. The channel of four CNN layers is 16, 32, 32, 64, the kernel size is 5, 3, 3, 3, and all the strides is 2. The encoder backbone of ResNet$_{-}$, ResNet$_{\oplus}$, and DUDA is ResNet-18~\cite{he2016deep}, which we directly use the implementation given by PyTorch without pretrained parameters. As for BCNN, we use the VGG-18~\cite{simonyan2014very} implemented by PyTorch as the backbone, which is to keep consistent with the original paper~\cite{lin2015bilinear}. In the decoder part, the output of the encoder is first flattened and then encoded by a fully-connected layer to become a 128-dimension vector. This 128-dimension vector will be sent to the adapted GRU network. In the GRU network, the hidden size of each GRU cell is 128 and two 1-layer fully-connected layers are used to decode the object vector and the value of each step respectively. The dimension of the object vector is 19 including 8 for the color, 3 for the size, 3 for the shape, 3 for the material, and 2 for the position. The dimension of the value output is 33.

\setlength{\tabcolsep}{5pt}
\begin{table}[t]
\begin{center}
\begin{small}
    \begin{tabular}{lrrrr}
    \toprule                
     Model            &   Encoder Backbone  &  Decoder             & Parameters  \\
    \midrule
    CNN$_{-}$         & 4-layer CNN         & Adapted GRU           & 737K \\
    CNN$_{\oplus}$    & 4-layer CNN         & Adapted GRU           & 738K \\
    ResNet$_{-}$      & resnet18            & Adapted GRU           & 11M \\
    ResNet$_{\oplus}$ & resnet18            & Adapted GRU           & 11M \\
    BCNN              & vgg11\_bn           & Adapted GRU           & 41M \\
    DUDA              & resnet18            & Adapted GRU           & 18M \\
    \bottomrule
    \end{tabular}
\end{small}
\end{center}
\caption{The architectural details of different baseline models under the TranceNet framework.}
\label{tab:model}
\end{table}
\setlength{\tabcolsep}{6pt}

The optimizer we used is Adam~\cite{DBLP:journals/corr/KingmaB14} and the learning rate is 0.001 in the beginning then reduced to 0.0001 after 25 epochs. The samples for Event and View is shared and the number of samples in the training, validation, and test set is 500,000, 10,000, and 20,000 respectively. Please note for each sample in View, we have 3 different final views so that the total image pairs are tripled. As for Basic, we collect all existing 1-step samples in data, and the size of training, validation, and test set is 125000, 2,500, and 5,000. All models are trained with 50 epochs on the training set and models that have the best results on the validation set are chosen to be evaluated on the test set to get the final results. In our experiments, images are resized to $120\times 160$ for fast training. Furthermore, by following the common practice on image augmentation~\cite{krizhevsky2012imagenet}, we apply a $0\sim5\%$ spatial translation to all input image pairs during training.

During evaluation, when moving an object from the visible area into the invisible area, any directions and steps that could cause the same effect without making objects overlapping are accepted. In the evaluation system, this is implemented by only comparing the visible objects' attribute values of the two final states, i.e.~the predicting final state and the ground truth final state.

\section{Human Test System} \label{sec:human_test}

\begin{figure*}[t]
    \centering
    \includegraphics[width=0.75\linewidth]{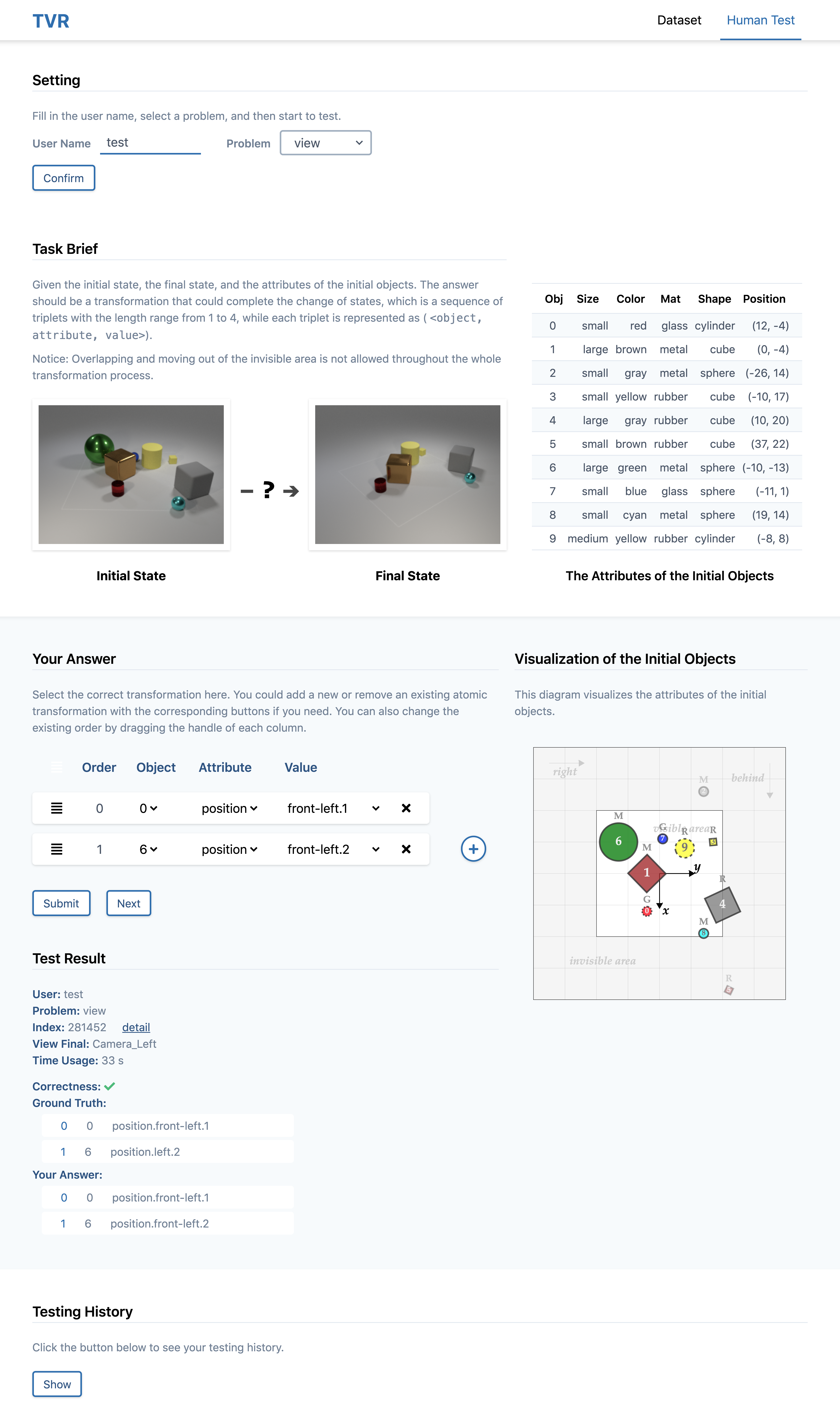}
    \caption{Human test system.}
    \label{fig:human_test}
\end{figure*}

To collect the human results, we build a web-based test system. Figure \ref{fig:human_test} shows the GUI of this system. The whole testing process includes the following steps. First of all, a human tester is told to be familiar with our system by trying a few examples with guidance. After that, the tester can start to test. During testing, for each sample, the tester should first observe given images and the attributes of the initial objects and then select the correct atomic transformations arranged with a feasible order. To reduce the time usage, we also provide the visualization of the initial objects for testers. After completing all samples, the tester can see his or her test result by checking the testing history.

\section{More Examples from TRANCE} \label{sec:examples}

The remaining pages show extra examples from the three settings of TRANCE, i.e.~Basic, Event, and View. In each sample, the initial state, the final state, and the attributes of the initial objects are provided. In the View setting, the view of the final state is randomly selected from Left and Right. To make readers easy to understand the given examples, for each example, an additional diagram is provided to visualize the attributes of the initial objects. At last, we show the reference transformation.

\begin{figure*}[t]
    \centering
    \includegraphics[width=0.95\linewidth]{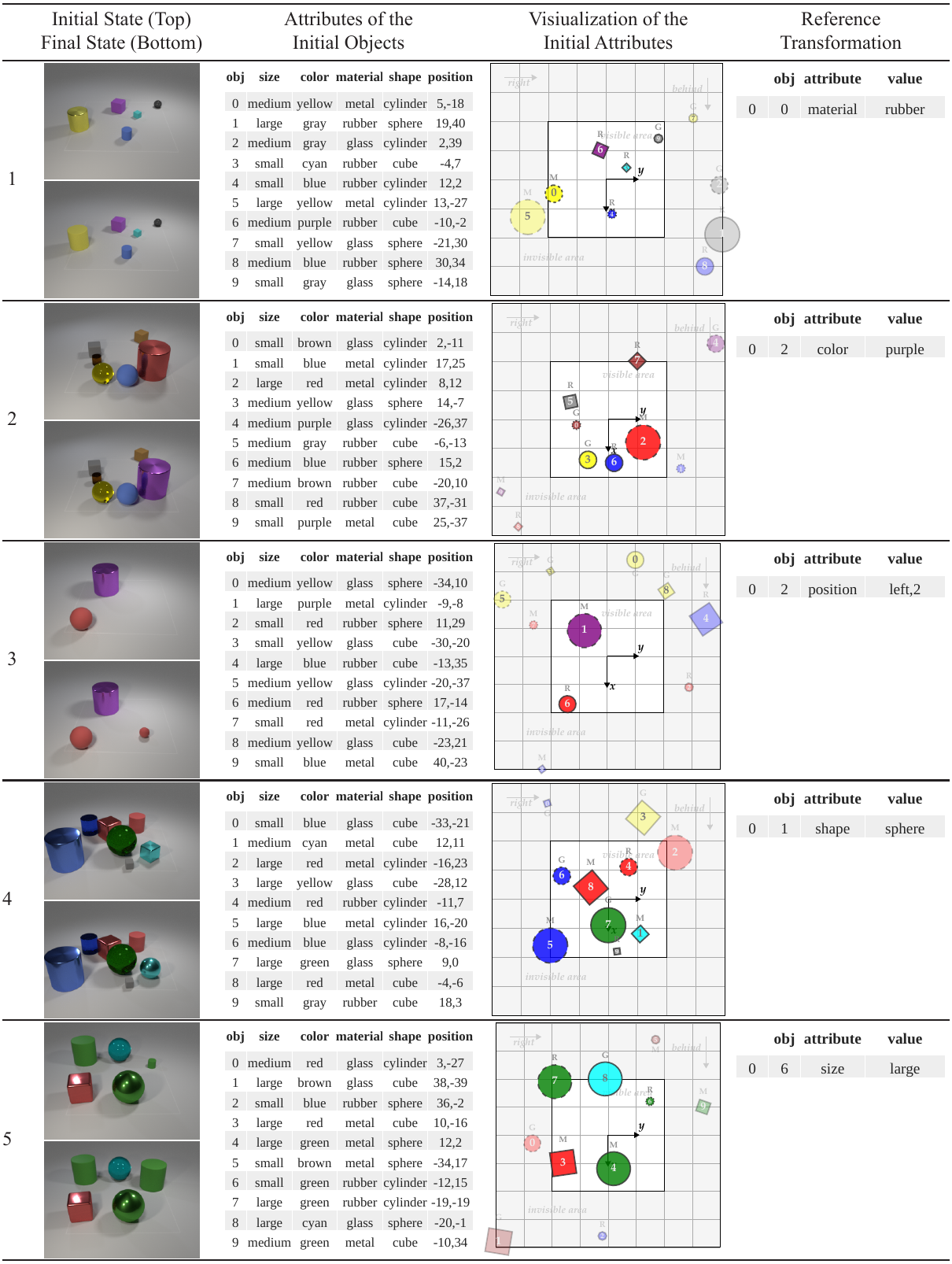}
    \caption{Examples from the Basic setting.}
    \label{fig:example_basic}
\end{figure*}

\begin{figure*}[t]
    \centering
    \includegraphics[width=0.95\linewidth]{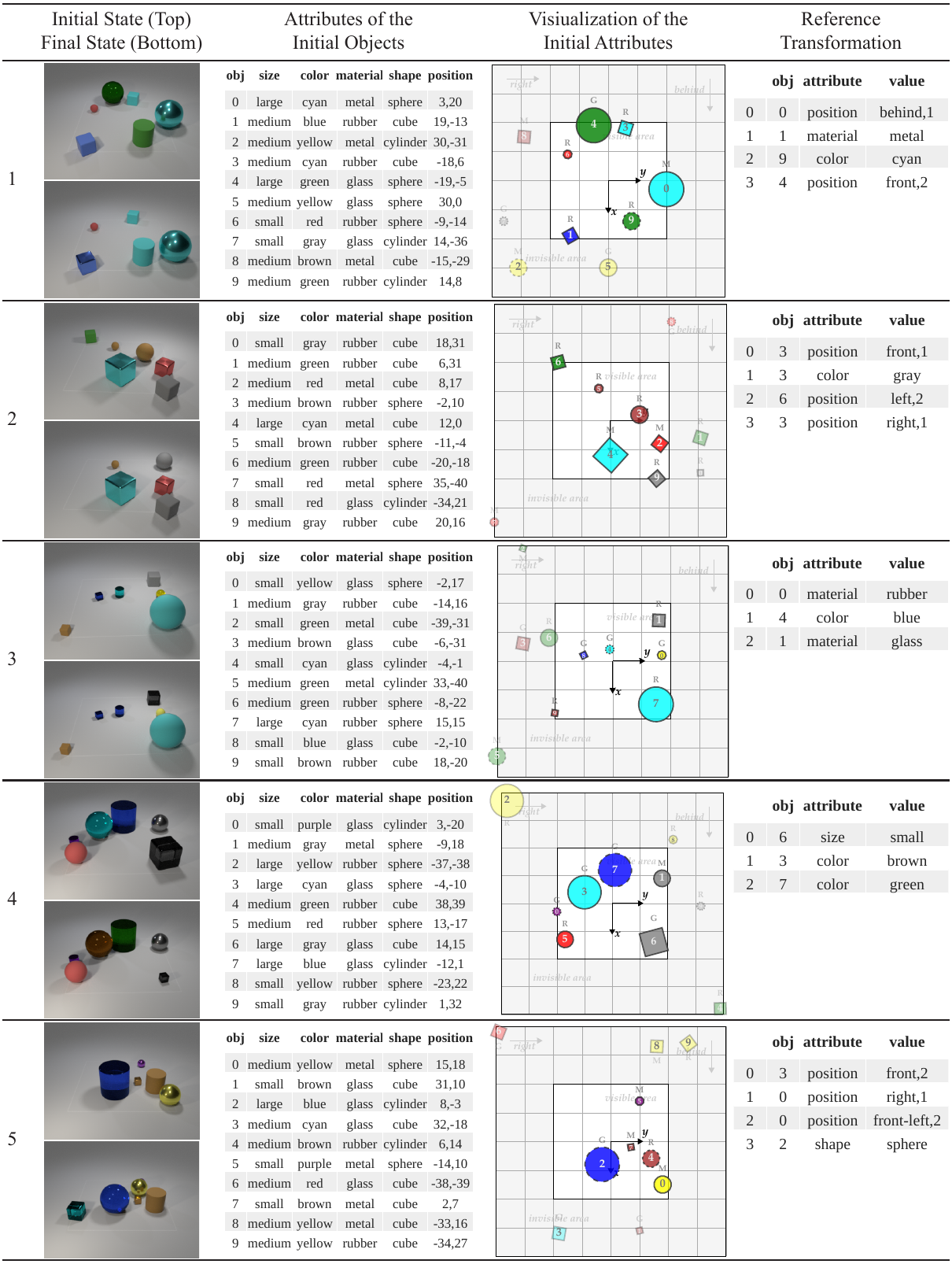}
    \caption{Examples from the Event setting.}
    \label{fig:example_event}
\end{figure*}

\begin{figure*}[t]
    \centering
    \includegraphics[width=0.95\linewidth]{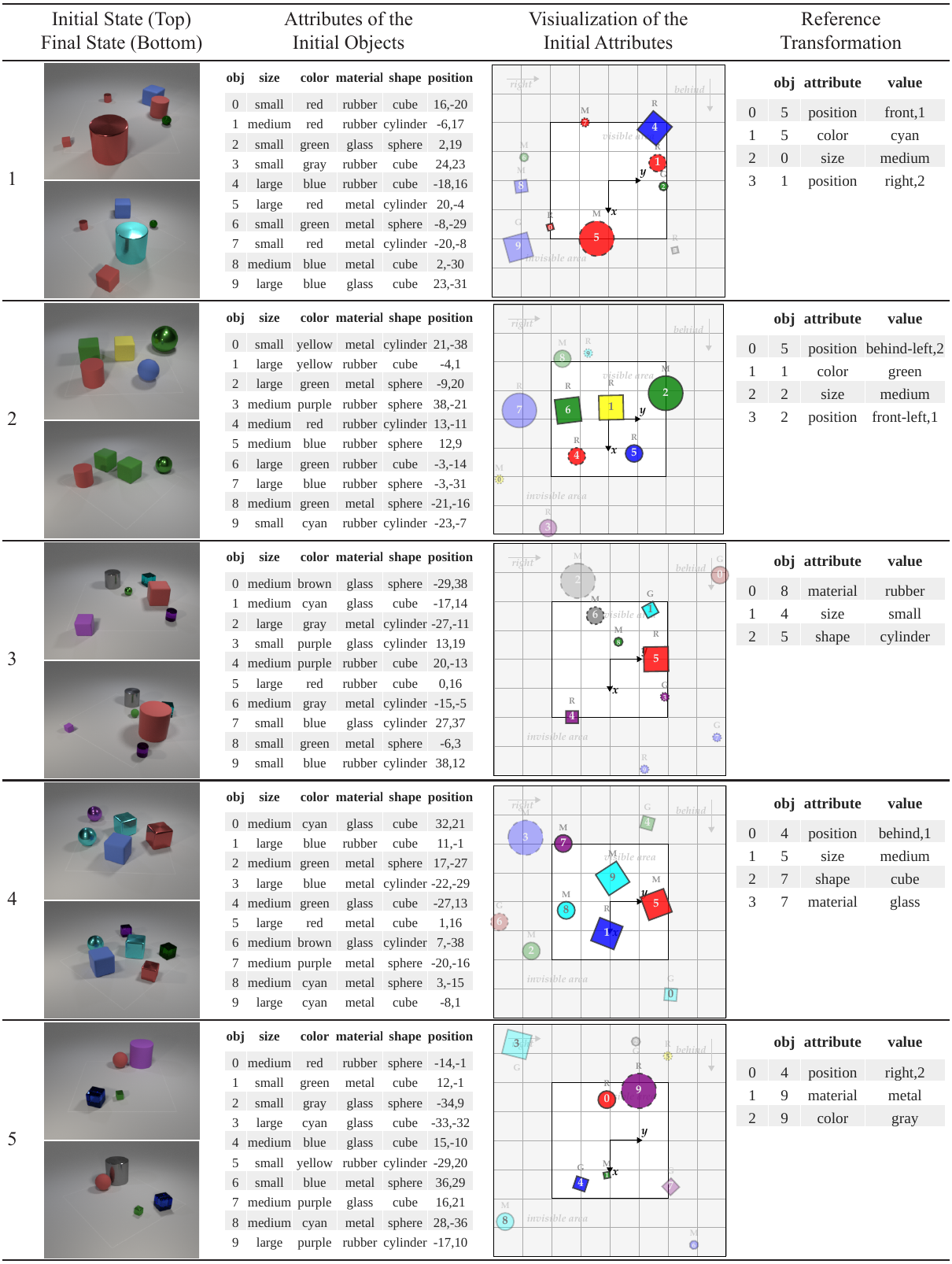}
    \caption{Examples from the View setting.}
    \label{fig:example_view}
\end{figure*}

\clearpage
\balance
{\small
\bibliographystyle{ieee_fullname}
\bibliography{main}
}

\end{document}